\documentclass[times,twocolumn,final]{elsarticle}
\newcommand{\update}[1]{\textcolor{black}{#1}}

\usepackage{cag}
\usepackage{framed,multirow}

\usepackage{amssymb}
\usepackage{latexsym}

\usepackage{url}
\usepackage{xcolor}
\usepackage{array}
\usepackage{booktabs}
\definecolor{newcolor}{rgb}{.8,.349,.1}

\usepackage[normalem]{ulem}

\usepackage{hyperref}

\usepackage[switch,pagewise]{lineno} 

\journal{Computers \& Graphics}

\begin{document}

\begin{frontmatter}

\title{Lightweight integration of 3D features to improve 2D image segmentation}

\author[1,4]{Olivier \snm{PRADELLE}\corref{cor1}}
\emailauthor{olivier.pradelle@hexagon.com}{Corresponding Author}
    
\author[1,3]{Raphaëlle \snm{CHAINE}}
\author[4]{David \snm{WENDLAND}}
\author[1,2,3]{Julie \snm{DIGNE}}

\address[1]{Laboratoire d'InfoRmatique en Image et Système d'information (LIRIS), Lyon, France}
\address[2]{CNRS}
\address[3]{Université Lyon 1, Lyon, France}
\address[4]{Technodigit, 14 Portes du Grand Lyon, Neyron, 01700, France}

\begin{abstract}
Scene understanding has made tremendous progress over the past few years, as data acquisition systems are now providing an increasing amount of data of various modalities (point cloud, depth, RGB...). However, this improvement comes at a large cost on computation resources and data annotation requirements.
To analyze geometric information and images jointly, many approaches rely on both a 2D loss and 3D loss, requiring not only 2D per pixel-labels but also 3D per-point labels. However, obtaining a 3D groundtruth is challenging, time-consuming and error-prone.
In this paper, we show that image segmentation can benefit from 3D geometric information without requiring a 3D groundtruth, by training the geometric feature extraction and the 2D segmentation network jointly, in an end-to-end fashion, using only the 2D segmentation loss.
Our method starts by extracting a map of 3D features directly from a provided
point cloud by using a lightweight 3D neural network. The 3D feature map, merged with the RGB image, is then used as an input to a classical image segmentation network.
Our method can be applied to many 2D segmentation networks, improving significantly their performance with only a marginal network weight increase and light input dataset requirements, since no 3D groundtruth is required.

\end{abstract}

\begin{keyword}
\KWD 2D segmentation \sep Deep learning\sep Geometry \sep Multimodal
\end{keyword}

\end{frontmatter}

\section{Introduction}
Today's 3D \update{LiDAR} scanners are often equipped with cameras acquiring RGB pictures alongside \update{a point cloud} : 2D images provide colors and texture of the objects, while 3D data provides geometric relationships between objects in the scene, beyond their differences in color and texture.
Hence, adequately combining both types of data can leverage their respective advantages and help overcome their limitations (3DMV~\citep{dai20183dmv}, BPNet~\citep{hu2021bidirectional}). 
While initial scene datasets provided only RGBD information and 2D groundtruth data (NYU-V2 dataset~\citep{silbermanECCV12nyu}), many recent datasets (ScanNet ~\citep{dai2017scannet},2D-3DS~\citep{armeni2017dataset}) provide RGBD images and depth-reconstructed point clouds.
\update{The KITTI-360 dataset \citep{liao2021kitti360} directly provides 3D data laser scans and RGB images captured dynamically.}

More importantly \update{these datasets} also come with both 2D and 3D groundtruth labels, these datasets being usually prepared and labelled manually.
For the ScanNet dataset, the 3D instance segmentation was crowd-sourced to more than 500 coworkers~\citep{dai2017scannet}, using a specifically designed labelling interface, with CAD model alignments. 
This shows that the 3D groundtruth labelling task is highly non trivial to set up and work-intensive.
On the contrary labelling 2D images is much simpler, as it can rely on traditional image segmentation techniques with a user possibly merging or correcting the segmented parts and naming them.

In this context, we introduce a segmentation method exploiting both 2D and 3D information, able to remove all dependency on a 3D groundtruth, only relying on 2D labels. A lightweight encoder extracts features from a 3D point set which are used to improve 2D segmentation results.
\update{Since it works on a single view, our method can also be applied to RGBD data by trivially reconstructing a point cloud from the single view depth by discarding pixels with no depth information.}
\footnote{Our code is available at https://github.com/OPradelle/2DGuidedLight3D.}

\begin{figure}[!h]
\centering
\includegraphics[width=7.5cm]{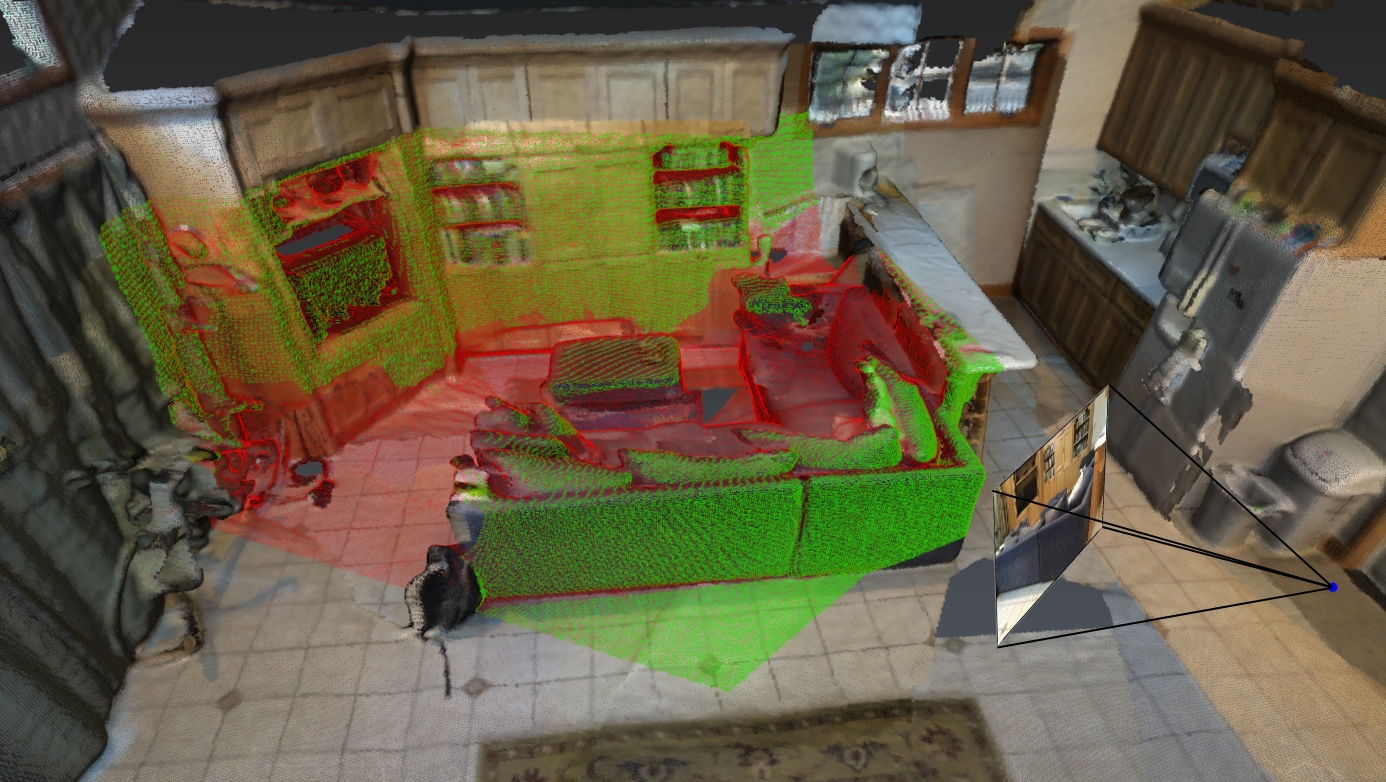}
\caption{Taking an RGB image \update{and} the points falling inside the viewing cone of the camera (red \update{and green} points), our method encodes geometric information and projects the visible points (green points) to the image plane before processing it with a 2D segmentation network.}
\label{fig:snapshot}
\end{figure}

\section{Related Work}
\paragraph*{2D image segmentation}
2D semantic segmentation has known several improvements over the last decade.
Long et al. ~\citep{long2015fcn} replaced the last layers of a convolutional classification network \citep{alexnet} with fully connected layers to produce per pixel segmentation. Ronneberger et al.~\citep{ronneberger2015unet} alternatively proposed U-Net, an architecture that shares information between coarsest and finest representation layers, making it more robust to missing groundtruth data.
Some approaches exploited even deeper networks ~\citep{szegedy2014googlenet} while others worked on the receptive fields of the convolution operation in order to achieve better context understanding. In particular, Yu and Koltun ~\citep{yu2015dilated} and Chen et al. ~\citep{chen2017dpl} used dilated convolution (also called atrous convolution) to gain contextual information. Dai et al. \citep{dai2017defconv} introduced a deformable convolution to allow the network to be more accurate in the area of interest. 
Recent approaches by Xie et al. ~\citep{xie2021segformer}, or Liu et al. ~\citep{liu2021Swin} introduce attention mechanisms to learn long range dependencies in the images for segmentation tasks.

\paragraph*{Deep Learning for 3D point cloud segmentation}
The main challenge of such sparse and unstructured data lies in the definition of a convolution operator working on points' neighborhoods, robust to point permutation, sampling changes and geometric transformations.
To overcome the lack of structure and define a convolution operation in 3D, some methods discretize the point cloud on a voxel grid and use 3D convolutional neural networks ~\citep{choy2019mink,liu2019pcnn} on such grids, while others rely on a nearest neighbor graph \citep{wang2018dgcnn} or use ball neighborhood queries \citep{thomas2019kpconv}.

To alleviate the need for a well-defined neighborhood, PointNet~\citep{qi2016ptnet} relies on per point convolutions with shared weights and symmetric aggregation operators 
to work on the raw point cloud directly. The network is then robust to sampling changes and point permutation, but at the cost of losing the locality of the computations. PointNet++~\citep{qi2017ptnet++} re-introduced some locality to the network, by using a multiscale computation. 
Other specific convolutions include defining the convolution weights as a continuous function on the local coordinates of 3D points \citep{wu2018ptconv}, or centering a convolution kernel around each point and defining the kernel features by summing the point features lying in the kernel's domain \citep{hua2017ptwiseconv}. 
Another way of handling 3D data is by projecting it to several 2D grids in a multi-view setting. 
Boulch et al.~\citep{boulch2017snapnet} and Kundu et al.~\citep{kundu2020mv} use a reconstructed mesh and render it from a free viewpoint. The rendered image is fed to a 2D semantic segmentation network \update{and} the predicted labels are projected back to the 3D points and merged across views. However, the mesh reconstruction process can be a costly step for large scenes.

\paragraph*{Merging 2D and \update{2.5-}3D data for scene segmentation}
Combining 2D and \update{2.5-}3D data can improve scene segmentation results by overcoming the limitation of each type of data. 
Gupta et al.~\citep{gupta2014learning} encode depth with RGB images and feed it to a segmentation network. But depth maps are view-dependent, and cannot account for the whole geometry of the scene, because of occlusions.

Another way to add \update{2.5D} information to the 2D segmentation network is to weight the convolution operator by local depth adequately \citep{wang2018depthaware}. Similarly, Cao et al.~\citep{cao2021shapeconv} redefine a convolution operator for RGBD images to embed the shape variation in the image. 
CMX ~\citep{liu2022cmx} uses a transformer network for RGB and depth images, merging both modalities between each encoder layer and using it during the decoding step as residual information. 

\update{To add 3D information to 2D images,} Liu et al.~\citep{liu20213d2ddist} \update{use a 3d network to extract features from point clouds and use them as cues to }
train the 2D network to emulate those 3D features. This allows the 2D network to produce more informative features for the 2D segmentation.
A crucial question when merging 2D and 3D features lies in where this merging should take place. One can either feed a combination of 2D and 3D information to a network (\emph{early} merge), or combine features in deeper layers, such as the bottleneck (\emph{late} merge).
For 3D point cloud segmentation, Jaritz et al.~\citep{jaritz2019multiview}, Dai and Nießner~\citep{dai20183dmv} use 2D features extracted by a 2D network working on multiple views as additional features to enhance a point set before feeding it to a segmentation network. They showed that this early merge gives better result than a late merging strategy, as it improves the propagation of the information from the images to the 3D data. 

Merging information at several \update{layer}'s depths has also been explored.
Su et al.~\citep{su2018splatnet} reconstruct a point set by multiview stereo, encode it in a voxel grid through a 3D CNN and project the encoded point set on a 2D grid. Because of the multiview reconstruction, each 3D point projects on a pixel and a standard 2D convolution can then be applied to the merged features at several depth in the 2D segmentation network. This method produces good results for object part segmentation, but does not scale well to large scenes.
For more generic input point sets, BPNet~\citep{hu2021bidirectional} similarly encodes 2D pixel and 3D voxel data independently, then merge\update{s} the representations at several layer depths during the decoding process.
However, in that case, interleaving voxel with pixel representations is not so straightforward. Indeed, the area of a voxel projected in an image depends on its distance to the image, hence no exact correspondence can be found.

All these 2D-3D combined approaches require a large amount of memory during training, and many require a 3D groundtruth~\citep{liu20213d2ddist, jaritz2019multiview, dai20183dmv, su2018splatnet, hu2021bidirectional}.
We propose a lightweight method to tackle the 2D segmentation task, using both image and geometric information.

\section{Overview}

Given a dataset containing a 3D point cloud and 2D images, we propose a network architecture to segment these images into semantic classes. Since the point cloud may come from a LiDAR device or simply be reconstructed from one or more depth images, we chose to work on the single view case.

We use a lightweight image segmentation network whose 2D input is enhanced by the output of a 3D point set encoder only supervised by 2D images segmentation groundtruth.

To summarize, our main contributions are:
\begin{itemize}
  \item A simple framework that takes a single image registered with a 3D point cloud
 and produces the image semantic segmentation in an end-to-end manner, improving the segmentation performance of several state-of-the-art image classifiers.
  \item A method to extract per view geometric information as a 2D map, much more informative than depthmaps.
  \item A method to exploit 3D information without requiring any 3D groundtruth.
\end{itemize}

\begin{figure}[!h]
\centering
\begin{tabular}{@{ }>{\centering\arraybackslash}p{1.8cm}@{ }>{\centering\arraybackslash}p{1.8cm}@{ }>{\centering\arraybackslash}p{1.8cm}@{ }>{\centering\arraybackslash}p{1.8cm}}
\includegraphics[width=1.8cm]{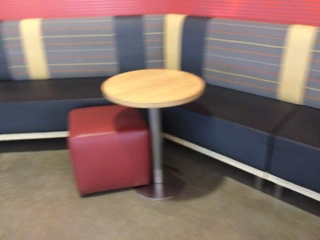}&
\includegraphics[width=1.8cm]{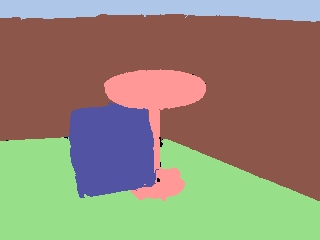}&
\includegraphics[width=1.8cm]{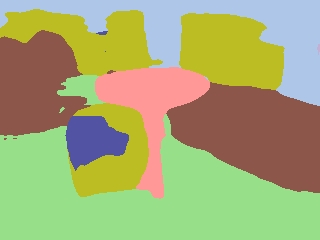}&
\includegraphics[width=1.8cm]{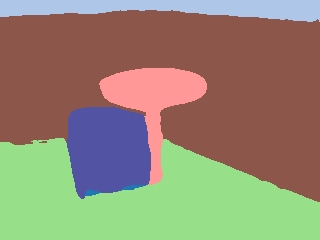}\\
\includegraphics[width=1.8cm]{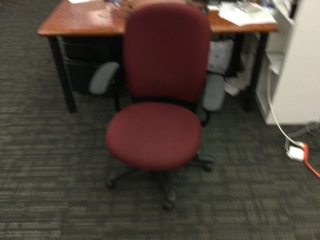}&
\includegraphics[width=1.8cm]{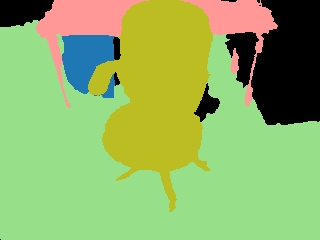}&
\includegraphics[width=1.8cm]{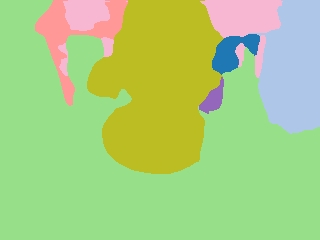}&
\includegraphics[width=1.8cm]{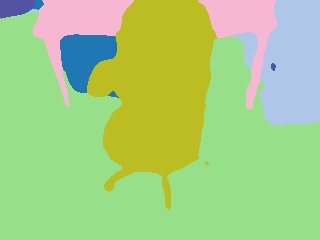}\\
\includegraphics[width=1.8cm]{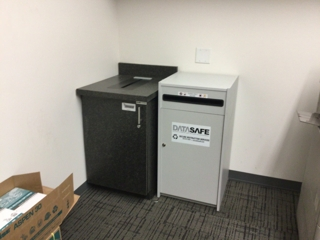}&
\includegraphics[width=1.8cm]{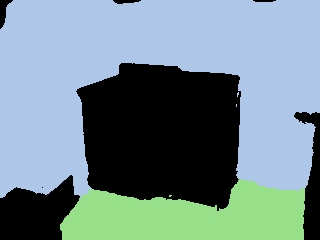}&
\includegraphics[width=1.8cm]{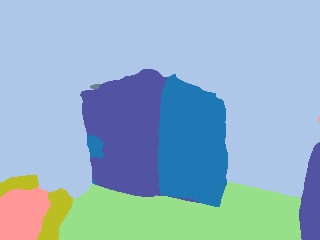}&
\includegraphics[width=1.8cm]{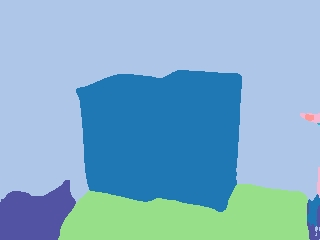}\\
\includegraphics[width=1.8cm]{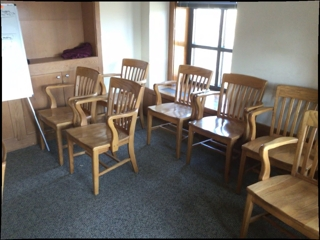}&
\includegraphics[width=1.8cm]{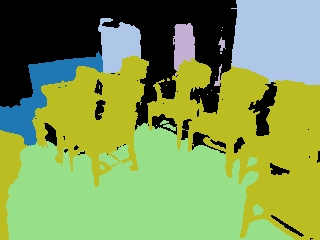}&
\includegraphics[width=1.8cm]{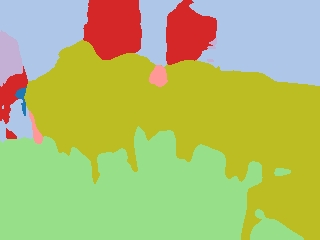}&
\includegraphics[width=1.8cm]{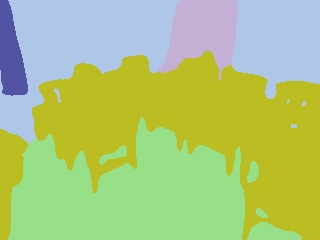}\\
\small{(a)RGB}&\small{(b)Groundtruth}&\small{(c)U-Net34}&\small{(d)U-Net34 + LPointNet}\\
\multicolumn{4}{p{1.8cm}}{\includegraphics[width=7.2cm]{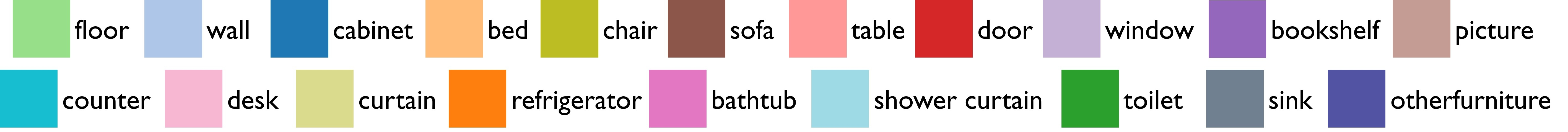}}\\
\end{tabular}
\caption{\update{Segmentation result on the ScanNet validation dataset. Merging the images with 3D information allows the network to predict more accurate object boundaries compared to the 2D baseline (U-Net34.)}}
\label{fig:segImg1}
\end{figure}

Figure \ref{fig:segImg1} illustrates some results of our method on the ScanNet \citep{dai2017scannet} dataset.


\section{Approach}

The goal of our method is to predict a semantic segmentation of a real-world RGB image with known pose and camera parameters by exploiting both RGB and 3D geometric information provided by an additional point cloud $P$.

First, a 3D encoder network processes the points located in the camera's viewing cone $P_{cone}$ and extracts 3D features for the set of visible points $P_{vis}\subset P_{cone}$. 

The advantage of having a point cloud of the scene, rather than just a depth image, is that the geometric information is not only computed from points in $P_{vis}$, but also from the points occluded in the view $P_{cone}$, as they give valuable context information (see the ablation study \ref{app:pvis}).
$P_{cone}$ and $P_{vis}$ are represented in Figure~\ref{fig:snapshot} with red and green points respectively.
We then project the 3D features from the $P_{vis}$ points onto the image plane using the camera parameters, creating a sparse 61-channel feature map. 
This feature map is view-dependent and can be thought of as an alternative to a depth map, encoding much richer geometric information. We then proceed to merge this feature map with the RGB image. 
The final step is to feed the combined image to a standard 2D image segmentation network to predict per pixel segmentation.

To train our network we use a cross entropy loss between the predicted labels and the ground truth image pixel labels. 
During training, we optimize not only the weights of the 2D segmentation network but also the weights 
of the 3D encoder network and the weights of the merging operation, guided solely by the 2D ground truth. This allows the 3D network to optimize the features for the 2D segmentation task.

\begin{figure}[!t]
\centering
\includegraphics[scale=0.5]{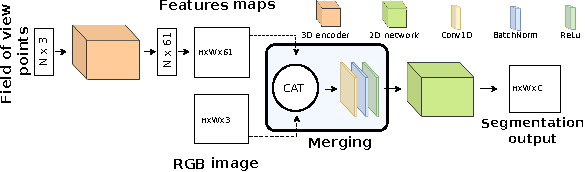}
\caption{The network takes as input a point cloud, an image and the corresponding projection matrix. The point cloud is processed using a 3D encoder to extract per visible point features that are projected on the image plane, yielding a 2D geometric feature map. 
The RGB image and the geometric feature map are merged 
using a single layer MLP before feeding it to a 2D segmentation network.}
\label{fig:network}
\end{figure}

The architecture of our network can be seen in Figure \ref{fig:network}. It outputs a label for every input pixel.

\subsection{Data preprocessing} \label{data}

Given an image and a point cloud $P$, let $P_{cone}$ be the scene's points that fall within the camera's viewing cone, and $P_{vis}\subset P_{cone}$ the set of points that are not occluded in the view. 

\begin{figure}[!h]
\centering
\begin{tabular}{@{ }>{\centering\arraybackslash}p{2.1cm}@{ }>{\centering\arraybackslash}p{2.1cm}
@{ }>{\centering\arraybackslash}p{2.1cm}}
\includegraphics[width=2.1cm]{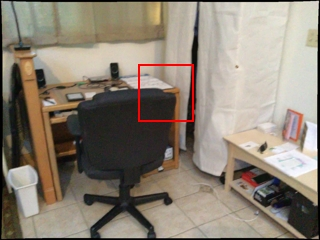}&
\includegraphics[width=2.1cm]{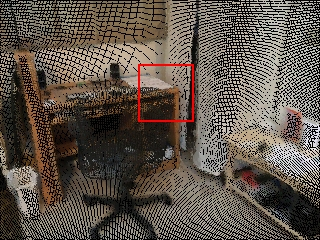}&
\includegraphics[width=2.1cm]{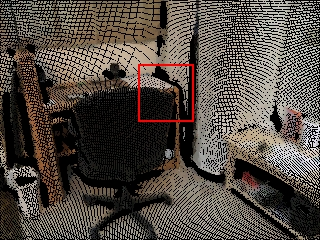}\\
\includegraphics[width=1.6cm]{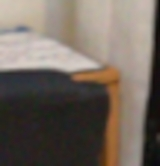}&
\includegraphics[width=1.6cm]{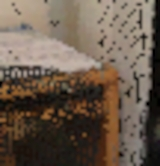}&
\includegraphics[width=1.6cm]{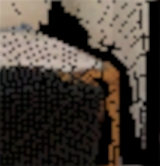}\\
\small{(a) RGB}& \small{(b) Closest projected points}&\small{(c) With visibility filter}\\
\end{tabular}
\caption{\update{Visibility filter effect.  Points of the table that are occluded by the chair are efficiently discarded. Pixels without geometric information are black. }}
\label{fig:visFilter}
\end{figure}

To construct the point set $P_{vis}$, we start by selecting, for each pixel, the point of $P_{cone}$ (if any) that is closest to the camera (Figure \ref{fig:visFilter}(b)).
Points that should be occluded by an object might still appear if the occluding object is too sparsely sampled (e.g. the desk behind the chair in Fig. \ref{fig:visFilter}).
To deal with these overlapping surfaces in sparsely sampled areas, we apply the visibility filter of Pintus et al.~\citep{pintus2011visible} (Figure \ref{fig:visFilter}(c)). After these steps, only few points remain per image.
In practice, only a low proportion of image pixels correspond to projected points  (only $15\%$ in the ScanNet dataset). 
Thus, the geometric information is very sparse.

At the end of the preprocessing, we obtain the camera's viewing cone points $P_{cone}$ and the selected subset of visible points $P_{vis}$.

In our framework, the 3D encoder uses all the points of $P_{cone}$ to compute features for a points of $P_{vis}$. Those features are then merged with the RGB image.

\subsection{Per view 3D feature map}

To compute geometric features, we choose to work directly on the raw point cloud to avoid any early discretization, as induced for example by a voxel grid, and to avoid costly preprocessing steps, such as mesh reconstruction. 

The only transformation is that we work on the $P_{cone}$ subset instead of the entire point cloud.

To extract 3D features, we can alternatively use  approaches that operate directly on raw point clouds such as PointNet~\citep{qi2016ptnet, qi2017ptnet++} and KPConv-CNN~\citep{thomas2019kpconv}.
However our experiments (section \ref{sec:xp}) show that KPConv-CNN is too memory-demanding due to the sheer number of points in $P_{cone}$, requiring to downsample the input point cloud. On the contrary, as PointNet shares its weights for all points, it is a very light network, which turned out to produce better features and is more consistent with our will to set up a lightweight structure. Hence we use PointNet by default.

\begin{figure}[!h]
\centering
\includegraphics[scale=0.5]{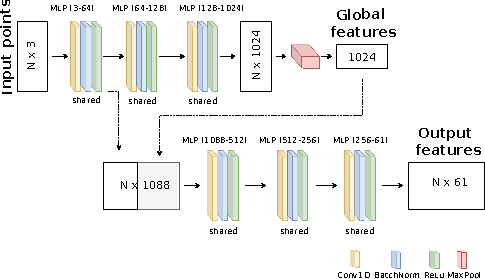}
\caption{LPointNet is a modified PointNet encoder. The spatial transformers are removed from the original architecture, as the points in $P_{cone}$ are expressed in the coordinate system of the camera.}
\label{fig:ptnet}
\end{figure}

We make a substantial change to the PointNet architecture: we remove the spatial transformers (t-nets) both in the 3D space and in the feature space.
Since we express the points coordinates with respect to the camera coordinate system, which is relevant in our case, we do not want to find an optimal global coordinate system for the points.
The relevance of this modification is demonstrated in our ablation study (Section \ref{ptnet}).
Furthermore, since these t-nets are in fact reduced PointNet instances,
we obtain a lighter version of this architecture that can be seen in Figure \ref{fig:ptnet} and we refer to it as LPointNet (for light PointNet) in the remainder of the paper.

LPointNet is fed with all the points in $P_{cone}$ to produce a global vector at pooling. This vector is further concatenated with the intermediate representation of each point in $P_{vis}$.
The output of our view-specific 3D encoder is a 61-channel feature vector per point of $P_{vis}$ that we store in an image at the corresponding pixel.

In practice, we do not rely on any pretrained weights for the 3D backbone (LPointNet or KPConv) and train it from scratch in an end-to-end manner. The ablation study also assesses the impact of this choice.

\subsection{Combining the 3D features and RGB values}
We locally merge the feature map with the color values pixelwise by concatenating the channels from the features map and the RGB image. The resulting 64 channel image is then fed to a 1D convolution layer, with batchnorm and ReLU activation. In case no point projects on a pixel, we directly use a 1x1 convolution to map the 3-channel color information to 64 channels.
This yields a combined 64-channel image which is processed by a standard 2D image segmentation network to predict per pixel segmentation.
This merging step is shown as the layer between the 3D encoder and 2D encoder in Figure \ref{fig:network}.

\subsection{2D backbone}

We use standard networks for the 2D segmentation neural network: U-Net~\citep{ronneberger2015unet} (with ResNet-34~\citep{he2015resnet}, ResNet-50~\citep{he2015resnet} or ResNet-101~\citep{he2015resnet} encoders), DeeplabV3~\citep{chen2017dpl} or SegFormer~\citep{xie2021segformer}. Among all these variant, we favor ResNet-34 which is lighter and well performing (see Table \ref{tab:var2D}).
The only difference with the \emph{vanilla} versions of these networks is that they are adapted to take 64-channel images instead of classical RGB images as input.
In practice, the 2D encoders were pretrained on the ImageNet dataset~\citep{russakovsky2015imgnet}, with the exception of the first convolution layer, since it takes a 64-channel input instead of a 3-channel input. These first layer weights are initialized randomly.


\section{Training} \label{train}

We \update{trained and evaluated} our segmentation method \update{on} 3 indoor datasets providing 2D images and geometric information either as a point cloud (ScanNet~\citep{dai2017scannet}, 2D-3D-S~\citep{armeni2017dataset})) or as separate depth information for all views (NYU-V2~\citep{silbermanECCV12nyu}).

For the ScanNet dataset, we follow the setup described in BPNet~\citep{hu2021bidirectional} and MVPtnet~\citep{jaritz2019multiview} which uses an image resolution of 320x240 during the training.
For the NYU-V2 and 2D-3D-S, we follow the setup from CMX~\citep{liu2022cmx} and use an image resolution of 640x480 and 480x480 respectively at train time and test time.

\update{For each dataset, we follow the train and validation split proposed by the original authors. ScanNet and 2D-3D-S datasets define the split depending on the scene digitized, while the NYU-V2 provide a per image split with no overlap between image to avoid the risk of overfitting.}

All these datasets provide the camera's intrinsic parameters, and its various poses along with the point cloud. They also provide groundtruth per pixel labels.
More details on the preprocessing steps of these datasets can be found in \ref{app:dataset}.

The network was trained with a Stochastic Gradient Descent (SGD) optimizer for $40$ epochs. To stabilize the learning, we divide the learning rate by $2$ every $5$ epochs, which we have found to give satisfactory results.
We use an initial learning rate of $0.01$ with a weight decay of $0.0001$ and a momentum of $0.9$.
%


\section{Experiments}
\label{sec:xp}

All our experiments were run on a computer with an Nvidia RTX quadro 6000 GPU.

\begin{table*}[!htb]
\caption{Comparison on the ScanNet validation set, with state-of-the-art single view methods. Methods with a * indicates that we retrained the networks using the code given by the original authors.}
\centering
\begin{tabular}{p{5.2cm}|>{\centering\arraybackslash}p{4.5cm}|>{\centering\arraybackslash}p{1.cm}|>{\centering\arraybackslash}p{1.5cm}|p{2.3cm}|>{\centering\arraybackslash}p{1.cm}}
\toprule
  Methods  & InputType & GT & NbParam & 2D backbone & mIoU \\
\midrule
  CMX*~\citep{liu2022cmx}& RGB + Depth (HHA) & 2D & 66 M & SegFormer-B2 &  51.3 \\
  RFBNet~\citep{deng2019rfbnet} & RGB + Depth (HHA) & 2D & No info & ResNet-50 & 62.6 \\
  \textbf{Ours (LPointNet + U-Net34)}& RGB + Point cloud from Depth & 2D & 26 M & ResNet-34 &  \textbf{63.2} \\
  SSMA~\citep{Valada_2019ssma} & RGB + Depth (HHA)& 2D &  56 M & AdaptNet++ & 66.3 \\
  ShapeConv~\citep{cao2021shapeconv} & RGB + Depth (HHA) & 2D & 58 M & Deeplabv3+ & 66.6 \\
\midrule
  3D-to-2D distil \cite{liu20213d2ddist} & RGB + Point cloud & 2D & 66M & ResNet-50 & 58.2 \\
  \textbf{Ours (KPConv + U-Net34)}& RGB + Point cloud & 2D & 49 M & ResNet-34 & \textbf{63.8}\\
  BPNet*~\citep{hu2021bidirectional}& RGB + Point cloud & 2D/3D & 96 M & ResNet-34 & 64.4 \\
  \textbf{Ours (LPointNet + U-Net34)}& RGB + Point cloud & 2D & \textbf{26 M} & ResNet-34 & \textbf{66.1}  \\
  VirtualMVFusion~\citep{kundu2020mv} (single view) & RGB + Normals + Coordinates & 3D & No info & xcpetion65 & 67.0 \\
  \textbf{\underline{Ours (LPointNet + SegFormer-B2)}}& RGB + Point cloud & 2D &  30 M & SegFormer-B2 &  \textbf{\underline{69.0}} \\
\bottomrule
\end{tabular}
\label{tab:scannetVal}
\end{table*}

\subsection{ScanNet validation dataset}
We tested our method on the validation set of the ScanNet dataset~\citep{dai2017scannet} and compare our results with state of the art methods (Table \ref{tab:scannetVal}).
These methods may not take the same kind of input as our approach.

To fairly compare with methods that take depth images as input, we also made experiments where we fed our networks with a point cloud generated from the \update{single view} raw depth information \update{with no occluded points}.

The scores for SSMA~\citep{Valada_2019ssma}, RFBNet~\citep{deng2019rfbnet} and VirtualMVFusion~\citep{kundu2020mv} were taken from the papers, as they do not provide the  network weights nor the implementation. We reproduced the result given by 3D-to-2D distil using the code and weight available.
CMX does not provide results on the ScanNet validation set, but the authors provide the code and weights for this dataset. Using these, we obtained a mIoU score of only 51.3 on the validation set, even after retraining the network by following the procedure given by the authors.
However, CMX still performs well on the test set, with a mIoU score of 61.3, which is the most efficient among the RGBD methods. 
Table \ref{tab:scannetVal} the number of parameters of each method, taken from the official github repositories when they are available.

The first part of Table \ref{tab:scannetVal} compares methods working on RGBD or depth generated point clouds. 
In this setting, our approach gives better results than single view RGBD based method RFBNet~\citep{deng2019rfbnet}, even with a lighter 2D encoder (ResNet34 in our case compared to ResNet50 for RFB-Net). This indicates that combining 3D and 2D features is more powerful than 2.5D features, even without explicit 3D supervision.
SSMA~\citep{Valada_2019ssma} gets better result than our method on the validation set for RGBD methods. This performance comes from their training procedure : they first train two dedicated 2D segmentation network on depth and RGB images on the ScanNet dataset, before running a second training where the network learns to integrate the new features and then run a third training to best fit the segmentation head.
Nevertheless, when taking the scene point cloud, our approach exhibits even better segmentation results ($+4\%$), since it gives a finer geometric context which helps the network to extract meaningful 3D features.
ShapeConv~\citep{cao2021shapeconv} shows the best result on RGBD data. This result comes from the voting procedure, which aggregates multiple scale prediction. In addition, ShapeConv is twice as big as ours.

Comparisons with other methods closer to ours (using point cloud and images) are represented on the second part of Table \ref{tab:scannetVal}. All of these methods require camera's intrinsic and pose either to create ground truth (such as BPNet~\citep{kundu2020mv}) or to create pixel-points correspondence at training time to exchange information between 2D and 3D data(BPNet~\citep{hu2021bidirectional}, 3D-to-2D distil~\citep{liu20213d2ddist}).
To achieve a high performance, VirtualMVFusion~\citep{kundu2020mv} requires a very specific and heavy data preparation step, where the scene is reconstructed by multiview stereo and mesh reconstruction allowing to render new views during training. On the contrary our lightweight modification of a 2D backbone (SegFormer-B2 in that case) yields better performances with a much lighter data preprocessing step.

In the case of BPNet~\citep{hu2021bidirectional}, the final prediction is obtained by aggregating multiple predictions for an image. We trained the network provided by the official github repository on 2 GPU for the comparison to be meaningful.
Compared to BPNet, our method does not require 3D groundtruth. It also avoids any discretization step, since we process the point cloud directly instead of a set of voxels.

Using a stronger 2D backbone, such as Segformer-B2 further pushes our network performance at a reasonable cost (30M of parameters). However, since the images need to have a higher resolution (640x480), the training time almost doubles for a gain of $2.9\%$ compared to the gain obtained on the U-net34 architecture.

Adding either 3D backbones (LPointNet or KP-Conv) to a standard 2D backbone still keeps the number of parameters below other state of the art methods. It is particularly visible on the methods using point clouds and images, with our method having at least twice less parameters. As a result, our approach requires less memory, both during training and at inference time while being as or more efficient than the state of the art methods.
The proposed method can thus run on a single low-end GPU. Table \ref{tab:memScannet} presents the number of parameters and the training times of some of the most efficient methods, with available codes, on the ScanNet dataset. The number of parameters was obtained using the official repositories on github.

\begin{table}[!h]
\caption{Memory consumption of the training step on the ScanNet dataset. We could not report the information for Virtual MV fusion since there is no official code available.}
\centering
\begin{tabular}{p{2cm}|>{\centering\arraybackslash}p{1cm}|p{3.cm}|p{1.3cm}}
\toprule
  Methods  & Nb Param & GPU Memory & Training time \\
\midrule
  CMX & 66M & 2080Ti (4x11Gb) & 3 days \\
  BPNet & 96M & RTX6000 (4x48Gb) & 2 days \\
  Ours(U-Net34) & \textbf{26M} & RTXQuadro6000 (24Gb) & \update{\textbf{1.5 days}}   \\
\bottomrule
\end{tabular}
\label{tab:memScannet}
\end{table}

\begin{figure}[!htb]
\centering
\begin{tabular}{@{ }>{\centering\arraybackslash}p{1.8cm}@{ }>{\centering\arraybackslash}p{1.8cm}@{ }>{\centering\arraybackslash}p{1.8cm}@{ }>{\centering\arraybackslash}p{1.8cm}}
\includegraphics[width=1.8cm]{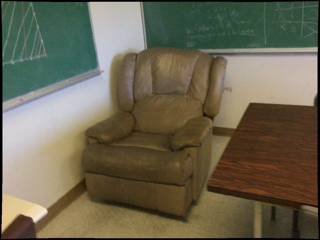}&
\includegraphics[width=1.8cm]{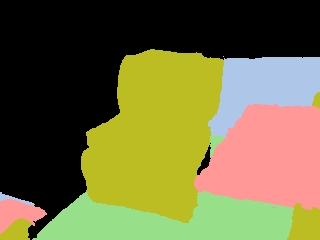}&
\includegraphics[width=1.8cm]{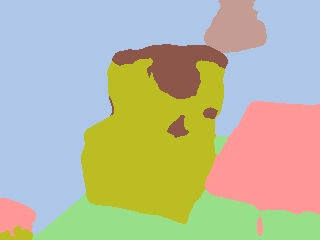}&
\includegraphics[width=1.8cm]{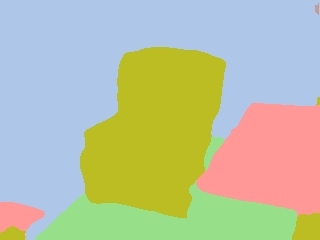}\\
\includegraphics[width=1.8cm]{image/seg/scene0046_00_000700.png}&
\includegraphics[width=1.8cm]{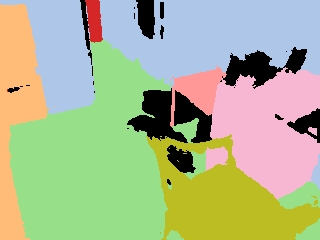}&
\includegraphics[width=1.8cm]{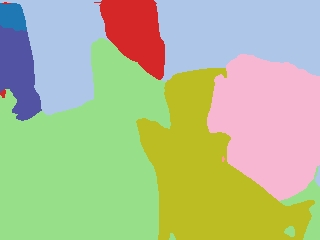}&
\includegraphics[width=1.8cm]{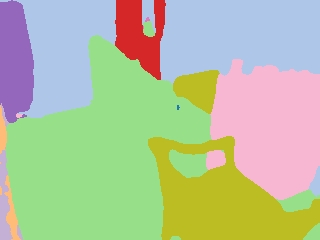}\\
\includegraphics[width=1.8cm]{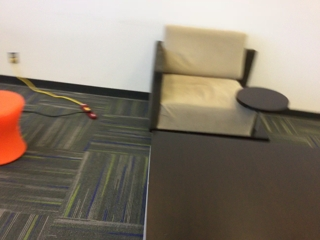}&
\includegraphics[width=1.8cm]{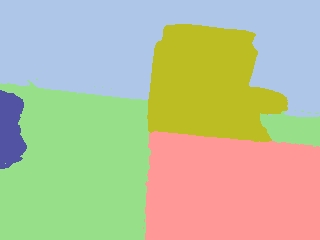}&
\includegraphics[width=1.8cm]{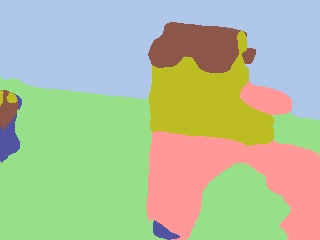}&
\includegraphics[width=1.8cm]{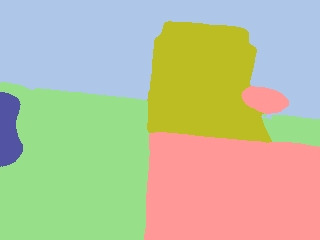}\\
\includegraphics[width=1.8cm]{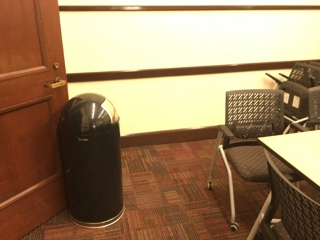}&
\includegraphics[width=1.8cm]{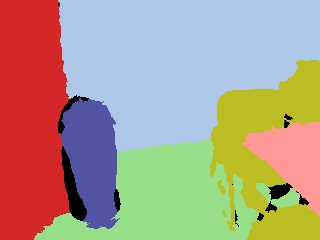}&
\includegraphics[width=1.8cm]{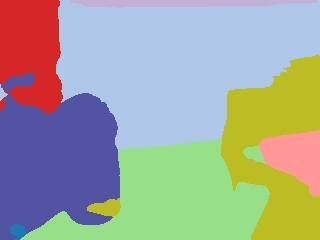}&
\includegraphics[width=1.8cm]{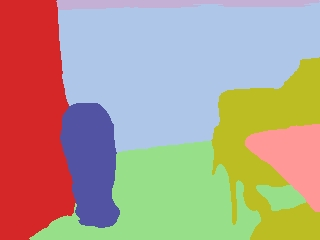}\\
(a)RGB&(b)GT&(c)U-Net34&(d)Ours\\
\multicolumn{4}{p{1.8cm}}{\includegraphics[width=7.2cm]{image/seg/legend.png}}\\
\end{tabular}
\caption{\update{Segmentation results on the ScanNet validation set}.}
\label{fig:segImg2}
\end{figure}

Figure \ref{fig:segImg2} illustrates the improvement of the segmentation results compared to the 2D baseline. As it can be seen on the image, the 3D features extracted from the point clouds allow the network to add consistency between the predicted pixel class and the object on the image. It also tends to better preserve the object's boundaries, as the 3D data allows to capture the object shape.

\paragraph*{3D features contribution}
To assess the contribution of our optimized 3D features, we compare our results with the same 2D backbone segmentation network applied on RGB, RGBD and RGBXYZ data.
RGBXYZ images correspond to 6-channels images concatenating RGB channels with the coordinates corresponding to points in $P_{vis}$, padded with $0$, if no point projects on the pixel.
Table \ref{tab:valset} shows that using RGBXYZ images increases the performance of the network compared to depth images. This indicates that the full spatial coordinates (in the camera coordinate system) information is more relevant than depth images for the task of 2D semantic segmentation, as it gives more information than simply the distance to the camera at each pixel.
Still, encoding the geometric information following our approach is far more efficient than using RGBDXYZ images, since adding 3D features significantly improves the segmentation (+10\% mIoU)

The fact that using our feature map instead of an RGBD image improves the segmentation by a large margin is particularly informative since depth images are often denser than our feature maps. 

Despite our sparser geometric information, the final segmentation is quantitatively improved, illustrating that our 3D feature extraction is more relevant than depth.

Finally, we experiment a different way of merging 2D and 3D information, by processing RGB and depth independently, using two dedicated U-Nets and merging the information at the bottleneck only.
Even compared to this configuration, our approach achieves significant improvement in terms of mIoU scores, with a very reasonable memory impact. Adding LPointNet to a U-Net approach only costs 1M parameters (26M parameters in total), while the dedicated U-Net almost doubles the total number of parameters (46M parameters). In all these experiments, we used a ResNet-34 encoder for U-Net, which provides good results while being comparatively small and fast to train.

\begin{table}[!htb]
\caption{2D mIoU on the validation set of the ScanNet dataset, for 2D U-Net segmentation networks based either on RGB images, RGBD images or for our method (* denote the addition of data augmentation).}
\centering
\begin{tabular}{p{5.5cm}|>{\centering\arraybackslash}p{0.8cm}|>{\centering\arraybackslash}p{0.7cm}}
  Methods  & mIoU  & mIoU gain\\
\midrule
  U-Net34 \citep{ronneberger2015unet} (RGB) & 55.5   \\
  U-Net34 (RGBD) & 59.3  & +3.8 \\
  U-Net34 (RGB + HHA) &  61.1 & +5.6  \\
  U-Net34 (RGB) + U-Net34 (Depth) & 61.2 & +5.7 \\
  U-Net34 (RGB) + U-Net34 (XYZ) & 62.3 & +6.8 \\
  \textbf{KPConv \citep{thomas2019kpconv} + U-Net34 (Ours)} & 63.8 & +8.3 \\
  \textbf{KPConv + U-Net34 (Ours)}* & 64.2 & +8.7 \\
  \textbf{LPointNet + U-Net34 (Ours)} & 65.4 & +9.9 \\
  \textbf{LPointNet + U-Net34 (Ours)}* & \textbf{66.1} & \textbf{+10.6} \\
\bottomrule
\end{tabular}
\label{tab:valset}
\end{table}

\subsection{2D-3D-S dataset}

Following the procedure recommended by the 2D-3D-S~\cite{armeni2017dataset} dataset, we use the images from Area\_5 as validation data, while the other areas are used to train the network.
The data was preprocessed following the approach described in appendix. As the 3D pointset covers the whole building floor and does not restrict to a single room, we use the provided depth map to filter out points belonging to other rooms, to reduce memory usage, as they are not relevant to the RGB image.
We used the same set of hyperparameters as for the ScanNet dataset during training.
We compare our 2D segmentation result with other methods on Table \ref{tab:2D-3D-S}.

\begin{table*}[!htb]
\caption{Comparison of the 2D-3D-S dataset. The first line of each block corresponds to the 2D backbone for the methods of the block (except for CMX, which does not provide results on their 2D backbone).}
\centering
\begin{tabular}{p{5.1cm}|>{\centering\arraybackslash}p{4.5cm}|>{\centering\arraybackslash}p{1.5cm}|>{\centering\arraybackslash}p{2.5cm}|>{\centering\arraybackslash}p{1.cm}}
  Methods  & InputType & NbParam & 2D baseline & mIoU \\
\midrule
  U-Net34~\citep{ronneberger2015unet} & RGB & 25M & ResNet-34 & 41.2\\
  SegFormer-B2~\citep{xie2021segformer} & RGB & 29M & SegFormer-B2 & 51.2 \\
  3D-to-2D distil~\cite{liu20213d2ddist} & RGB + Point cloud & 66M & ResNet-50  & 46.42 \\
 \textbf{U-Net34 + LPointNet (Ours)} & RGB + Point cloud & 26M & ResNet-34 & \textbf{53.5} \\
\midrule
  Deeplabv3+~\citep{chen2018dpl+}  & RGB + Depth (HHA) & 57M & ResNet-101 & 54.6\\
  CMX~\cite{liu2022cmx} & RGB + Depth (HHA) & 66M & SegFormer-B2 &  58.1 \\
 \textbf{Segformer-B2 + LPointNet (Ours)} & RGB + Point cloud from Depth &  \textbf{30M} & SegFormer-B2 &  \textbf{58.5}\\
  \underline{ShapeConv~\cite{cao2021shapeconv}} & RGB + Depth (HHA) & 58M & Deeplabv3+ & \underline{60.6} \\

\bottomrule
\end{tabular}
\label{tab:2D-3D-S}
\end{table*}

Compared to the 2D baseline (U-Net34 or SegFormer-B2), our approach improves the mIoU scores of the 2D baseline by more than $10\%$ for the U-Net34. This shows that using 3D data significantly helps the network to produce better segmentation result.
Among the methods which processes point clouds with RGB images, our approach outperforms 3D-to-2D distil \cite{liu20213d2ddist} even when using a smaller 2D architecture. Our improvement can be explained by the fact that we use real 3D data, while 3D-to-2D distil~\cite{liu20213d2ddist} only mimics 3D features.
We compare our approach with RGB-D state of the art method using a similar approach as the one used for the ScanNet~\citep{dai2017scannet} dataset. We use the depth map and the camera's parameter to create a per \update{single} view point cloud which is used as input for our approach.\\
CMX~\citep{liu2022cmx} yields a better score than ours, but at the price of a much heavier 2D backbone (SegFormer-B4~\citep{xie2021segformer}), with six times more parameters.
Using the SegFormer-B2 architecture, CMX achieves a mIoU score of 61.2.  At test time, CMX aggregate multiscale prediction to compute the final score.
To be fair with our approach, we trained CMX using the code provided by the authors on their github repositories and removed the multi-scale prediction at test time. In this configuration, CMX obtains a mIoU score of 58.1, while our approach reaches a score of 58.5. Our approach can thus improve the 2D network at a small cost using the geometric information without aggregating multiscale prediction. 
ShapeConv~\citep{cao2021shapeconv} is the best performing method on the 2D-3D-S dataset due to their training procedure : they apply heavy data augmentation technique and use the initial resolution of the image (1080x1080), allowing the network to capture finer details. However, this comes at a large training time cost compared to our approach.

From these observations, we can conclude that our approach offers competitive performances compared to RGBD approaches (Deeplabv3+, ResNet-101 on RGBD (HHA)) with half the number of parameters. However, replacing U-Net34 with new heavier \update{2D backbone }
(such as SegFormer) \update{in our architecture} yields even better results.

Several reasons explain the relative under-performance of our method on the 2D-3D-S dataset, where heavier architectures are more powerful. 
\update{Firstly, the image resolution taken during training makes the } \update{U-Net34 light 2D-backbone } 
\update{weaker compared to heavier architecture : methods with a larger receptive field, such as SegFormer, will give better result.}
\update{Secondly,} for this dataset, we down-sampled the \update{point set} to keep 60\% of the points. The feature map is thus sparser than the depth image used in ShapeConv~\citep{cao2021shapeconv} or CMX~\citep{liu2022cmx}. 
It is \update{even} sparser than the feature maps obtained in the ScanNet dataset. On average, 15\% of the pixels have a corresponding point on the ScanNet against 10\% \update{in our experiments with} 2D-3D-S.
However, compared to the 2D baseline, our approach obtains significant mIoU score gain, highlighting the 3D contribution for the image segmentation task.

\begin{figure}[!htb]
\centering
\begin{tabular}{@{ }>{\centering\arraybackslash}p{1.8cm}@{ }>{\centering\arraybackslash}p{1.8cm}@{ }>{\centering\arraybackslash}p{1.8cm}@{ }>{\centering\arraybackslash}p{1.8cm}}
\includegraphics[width=1.8cm]{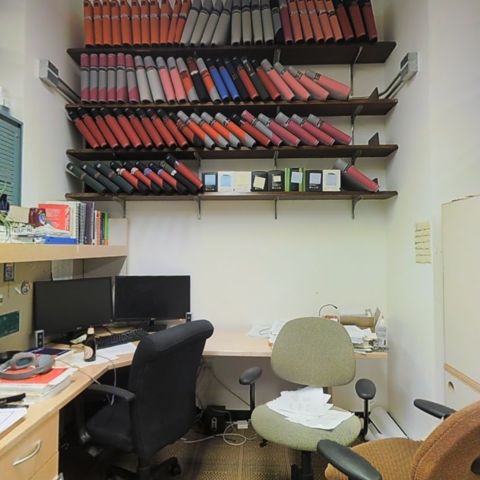}&
\includegraphics[width=1.8cm]{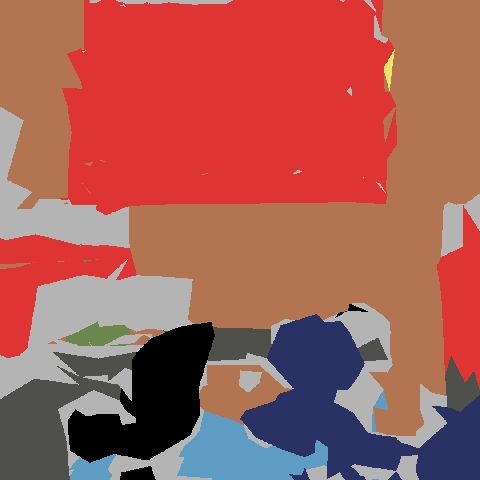}&
\includegraphics[width=1.8cm]{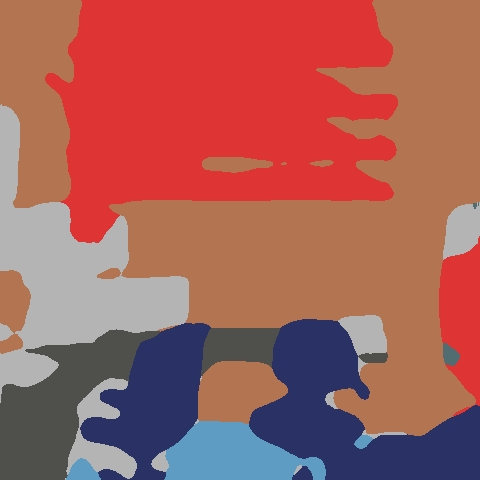}&
\includegraphics[width=1.8cm]{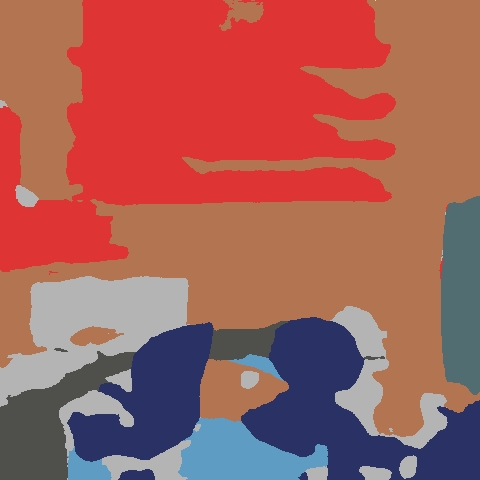}\\
\includegraphics[width=1.8cm]{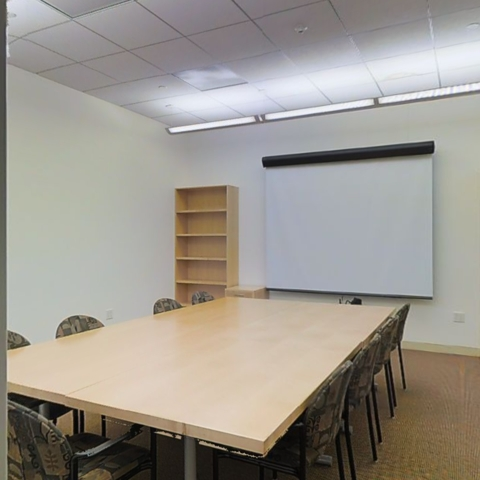}&
\includegraphics[width=1.8cm]{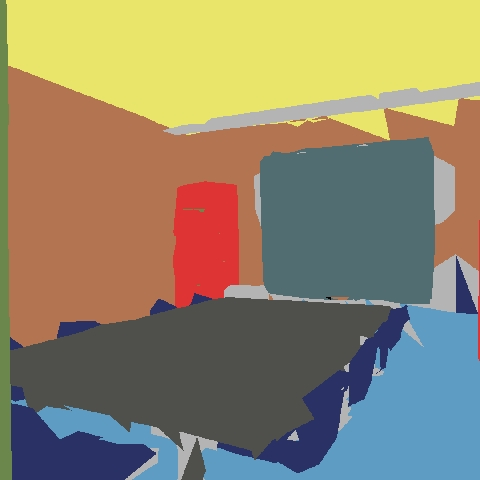}&
\includegraphics[width=1.8cm]{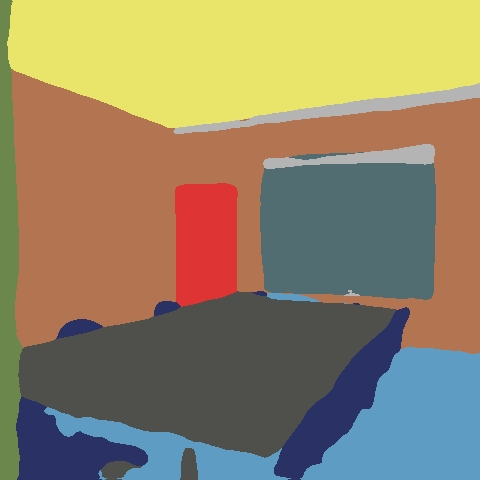}&
\includegraphics[width=1.8cm]{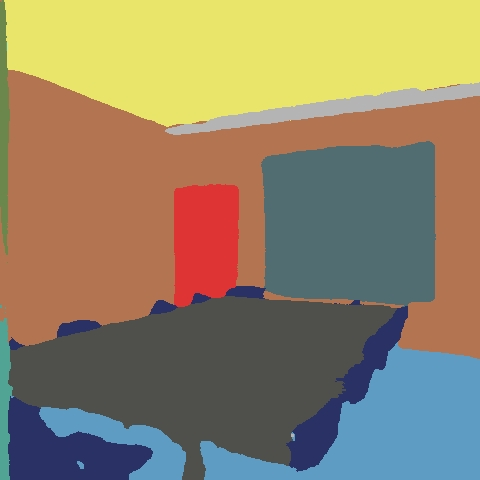}\\
\includegraphics[width=1.8cm]{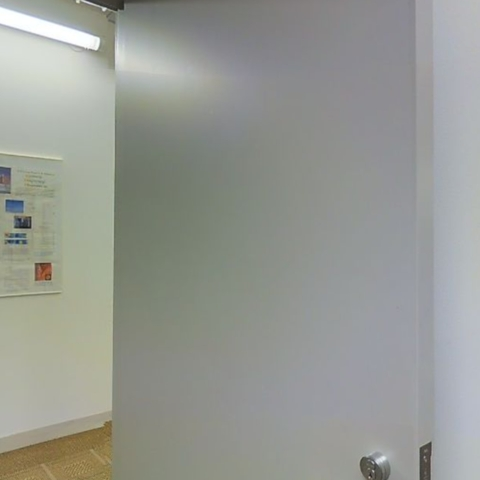}&
\includegraphics[width=1.8cm]{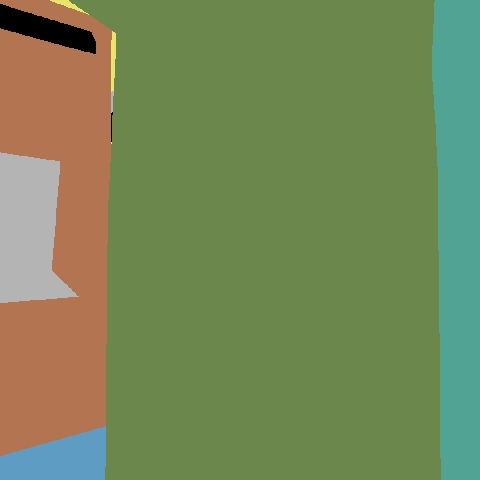}&
\includegraphics[width=1.8cm]{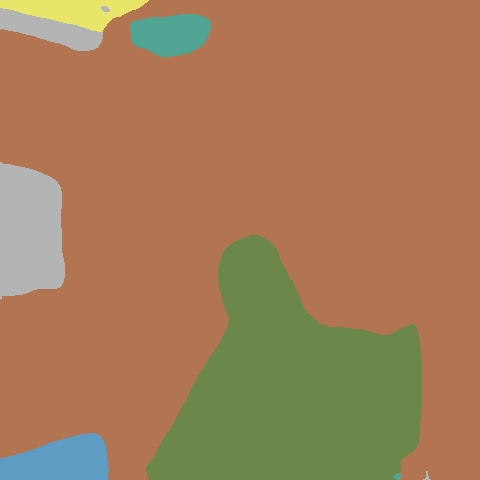}&
\includegraphics[width=1.8cm]{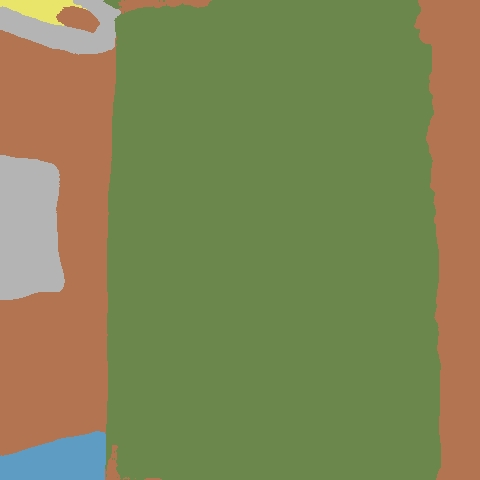}\\

(a)RGB&(b)GT&(c)SegFormB2&(d)Ours\\

\multicolumn{4}{p{2.1cm}}{\includegraphics[width=7.2cm]{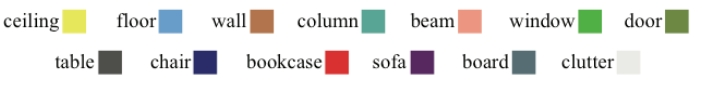}}\\
\end{tabular}
\caption{\update{Segmentation results on the 2D-3D-S validation set using SegFormer-B2 as 2D segmentation network.}}
\label{fig:segImg3}
\end{figure}

\update{Similarly to the U-Net34, a much stronger 2D backbone -such as the SegFormer-B2 architecture- also benefits from the 3D features encoded by the LPointNet network, as can be seen in Figure\ref{fig:segImg3}.}

\subsection{NYU-V2 dataset}\label{nyuv2}

In contrast with the other datasets, NYU-V2\citep{silbermanECCV12nyu} does not provide a standalone registered point cloud. But the dataset provides depth images either raw or in-painted and we can generate a point cloud for each camera using the camera's intrinsic parameter. We take the raw depth images given by the dataset, as the in-painted version produces a noisy cloud once projected. As a side effect of the per-camera point cloud, the visible point set of $P_{vis}$ directly corresponds to the depth map, and no visibility filter is needed. However, we no longer benefit from 3D information provided by hidden points and we can observe a drop in the quality of the results.

The NYU-V2 dataset provides 795 training images, and 654 validation of indoors scenes annotated with 40 classes. The results are reported in Table \ref{tab:valsetnyu}.

\begin{table}[!htb]
\caption{2D mIoU on the validation subset of the NYU-V2 dataset using single view.}
\centering
\begin{tabular}{p{7.cm}|>{\centering\arraybackslash}p{1.cm}}
  Methods  & mIoU \\
\midrule
  U-Net50 (RGB) &  30.5  \\
  \textbf{LPointNet + U-Net50 (Ours)} & 36.5  \\
\midrule
  SegFormer-B2 & 46.7 \\
  CMX(B2)~\citep{liu2022cmx} (Raw Depth) &  48.0\\
  CMX(B2)~\citep{liu2022cmx} (Raw Depth HHA) &  48.2 \\
  \textbf{\underline{ShapeConv~\citep{cao2021shapeconv} \update{(Inpainted Depth HHA)}}} &  \textbf{\underline{50.2}}  \\
  \textbf{\underline{SegFormer-B2 + LPointNet (Ours)}} &  \textbf{\underline{50.2}} \\
\bottomrule
\end{tabular}
\label{tab:valsetnyu}
\end{table}

Compared to the 2D baseline, adding depth point cloud information improves the segmentation result. However, the results are far less interesting than when considering a point cloud, as it allows the network to process more contextual information around the points.
Compared to ShapeConv~\citep{cao2021shapeconv} and CMX~\citep{liu2022cmx}, we used the raw depth coming from the sensor. For a fair comparison, we trained CMX~\citep{liu2022cmx} using the raw depth, with and without the HHA encoding, and replaced our 2D encoder by SegFormer-B2 \citep{xie2021segformer}. We used the code given by the authors and followed their training procedure using a single GPU. We removed the multi-scale prediction at testing time, to obtain the real performance of the network.
Table \ref{tab:valsetnyu} shows that our method gives better result than the CMX~\citep{liu2022cmx} method, highlighting our interest in the geometric data rather than the depth map. Our approach thus obtain similar result with ShapeConv~\citep{cao2021shapeconv}, with a lighter network and without relying on heavy data augmentation technique during training.

\subsection{Ablation study}
\label{ptnet}

To validate our architecture choices, we perform an ablation study using the validation set of the ScanNet~\citep{dai2017scannet} dataset, by changing the 3D point feature extraction network or the 2D network. We further test several modification of the PointNet architecture (see also the supplementary).

\paragraph*{Changing the 2D network}
Table \ref{tab:var2D} reports the performances obtained when using different 2D backbones.
Our experiments show that 3D feature information improves the performance of all U-Net34, U-Net50, U-Net101, SegFormer-B2 and DeeplabV3 networks. The performance improvement is particularly significant for U-Net34.
For DeeplabV3 it is less obvious because the network is efficient on images with a better resolution: the images used are 320x240 pixels, while DeeplabV3 preferentially uses 513x513 images during training, to alleviate the effect of image padding on dilated convolution.
In order to train the SegFormer architecture, we set the image size as 640x480 since the Transformer needs larger images to be efficient. This combination obtains the best score on the ScanNet~\citep{dai2017scannet} dataset, however it needs twice the time for the model to train for a light gain compared to the 2D baseline. Since LPointNet gives better results than KPConv, we choose to only test LPointNet with SegFormer-B2 and Deeplab as these methods take time to be trained and combining them with KPConv would be intractable.
These variations demonstrate the modularity of our approach on different kinds of architecture, and the contribution of the 3D features on the 2D segmentation task.

\begin{table}[!htb]
\caption{2D mIoU on the validation subset of the ScanNet dataset, for our method using ResNet-34, ResNet-50 encoder for U-Net, SegFormer-B2 or DeeplabV3 ResNet-50 architecture, using KP-Conv or LPointNet.}
\centering
\begin{tabular}{p{4.cm}|>{\centering\arraybackslash}p{1.2cm}|>{\centering\arraybackslash}p{0.8cm}|>{\centering\arraybackslash}p{1.cm}}
  Methods  & NbParam & mIoU & mIoU gain\\
\midrule
  U-Net34 & 25M & 55.5 & \\
  \textbf{LPointNet + U-Net-34} & 26M & 66.1 & \textbf{+10.6} \\ 
  KPConv + U-Net-34 & 49M & 64.2  & +8.7\\
\midrule
  U-Net50 & 86M & 58.2 \\  
  LPointNet + U-Net50 & 87M & 64.7  & +6.5 \\
  KPConv + U-Net50  & 110M & 64.0 & +5.8\\
\midrule
  U-Net101 & 105M & 58.4 \\  
  LPointNet + U-Net101 & 106M & 65.3  &  +6.9 \\
  KPConv + U-Net101 & 129M &  64.1 & +5.7\\ 
\midrule
  SegFormer-B2 & 29M & 65.8 \\  
  LPointNet + SegFormer-B2 & 30M & \textbf{69.0} & +3.2 \\
\midrule
  DeeplabV3 & 39M & 60.1 \\
  LPointNet + DeeplabV3 & 40M & 64.5 & +4.4\\
\bottomrule
\end{tabular}
\label{tab:var2D}
\end{table}

\paragraph*{Changing the 3D network}
Our approach preferentially relies on PointNet, but it can straightforwardly be adapted with a different 3D neural network. For example one can switch to KPConv-CNN~\citep{thomas2019kpconv} which uses a ball centered at a query point, a constant sample distribution in this ball and a special convolution on these samples. Since KPConv is memory intensive, we were only able to use a downsampled version of the ScanNet point cloud using Poisson sampling~\citep{Cook86} with a query ball of 3cm. On the validation dataset, we obtained a mIoU of \update{$64.2$} (see Table. \ref{tab:var2D}). 
We followed the architecture and data processing given by the KPConv authors without fine tuning it for our case. We set the starting learning rate at $0.001$ and add dropout before using the KPConv operation in each encoder layers.

Despite the need of a 3D subsampling, we observe that KPConv improves the performance of 2D segmentation network. However compared to the LPointNet variation, the number of parameters is twice as much (\update{49M} against 26M). The heavy memory consumption of KPConv goes against our search for a lightweight architecture and this is why we defaulted to LPointNet in all other tests.

\paragraph*{PointNet Architecture choices}
In our approach, we chose to remove the spatial transformers (t-net) from the original PointNet architecture as we observe the cloud from the camera point of view. 
This 3D orientation is relevant in the scene analysis case, since a floor is likely to be lower than a ceiling, and cannot be vertical. Moreover, it also reduces the weight of the Pointnet structure.
To validate this experimentally, we compared the performances with and without these t-nets, the network being retrained in each case. 
Table \ref{tab:ptnetarch} shows that using the t-nets degrades the mIoU scores. 

\begin{table}
\caption{Effect of using or removing spatial transformers (t-net) on the validation set of the ScanNet Dataset (* denote the addition of data augmentation).}
\label{tab:ptnetarch}
\centering
\begin{tabular}{p{7.cm}|>{\centering\arraybackslash}p{1.cm}}
  Methods  & mIoU \\
\midrule
  PointNet \citep{qi2016ptnet} + U-Net34 \citep{ronneberger2015unet}  &   62.8 \\
  PointNet (w/o spatial's t-net) + U-Net34 & 62.9 \\
  PointNet (w/o feature's t-net) + U-Net34 & 62.6 \\
  \textbf{LPointNet + U-Net34 (Ours)}  &  \textbf{65.4} \\
  \textbf{LPointNet + U-Net34 (Ours)}* & \textbf{66.1} \\
\bottomrule
\end{tabular}
\end{table}

\paragraph*{Pretraining LPointNet} We test whether pretraining LPointNet can improve the segmentation performance, using a network trained on point sets of 2D-3D-S for 3D semantic segmentation. Table \ref{tab:pretrained} shows that this does not improve the performance. It tends to suggest that the features needed for 3D segmentation might be slightly different than the ones used for helping a 2D segmentation.
In the same way, we test the effect of pretraining the 2D encoder on the ImageNet~\citep{russakovsky2015imgnet} Dataset.
We observe that using a pretrained 2D segmentation network provides a better starting point for the network, allowing it to extract more relevant features on combined feature maps.

\begin{table}[!htb]
\caption{Effect of using pretrained networks on the validation set of the ScanNet Dataset (* denote the addition of data augmentation).}
\label{tab:pretrained}
\centering
\begin{tabular}{p{7.cm}|>{\centering\arraybackslash}p{1.cm}}
  Methods  & mIoU \\
\midrule
  LPointNet + U-Net34 \citep{ronneberger2015unet}  &   62.5 \\
  LPointNet + U-Net34 (pretrained) (Ours)  &  65.4 \\
  \textbf{LPointNet + U-Net34 (pretrained) (Ours)} * & \textbf{66.1} \\
  LPointNet (pretrained) + U-Net34 (pretrained) &  64.7 \\
\bottomrule
\end{tabular}
\end{table}

Since the transformer based architectures need a lot of data and time to be trained efficiently from scratch, we did not test the effect of pretraining the SegFormer architecture by ourselves and directly use the pretrained model given by the authors (pretrained on ADE20K dataset~\citep{zhou2019ade20k}).

\subsection{LPointNet feature maps}
\begin{figure}[!htb]
\centering
\begin{tabular}{@{ }>{\centering\arraybackslash}p{2.1cm}@{ }>{\centering\arraybackslash}p{2.1cm}@{ }>{\centering\arraybackslash}p{2.1cm}@{ }>{\centering\arraybackslash}p{2.1cm}}
\includegraphics[width=2.1cm]{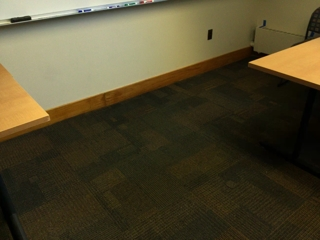}&
\includegraphics[width=2.1cm]{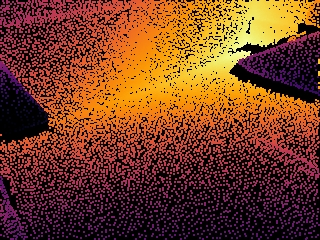}&
\includegraphics[width=2.1cm]{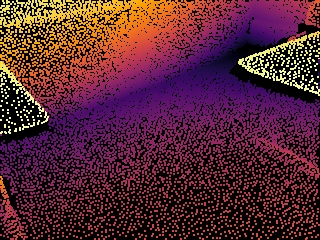}&
\includegraphics[width=2.1cm]{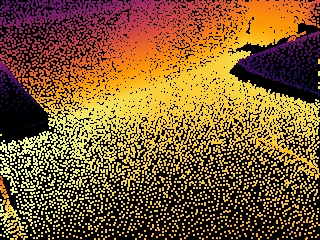}\\
\includegraphics[width=2.1cm]{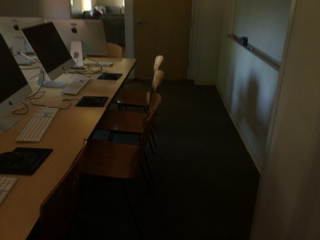}&
\includegraphics[width=2.1cm]{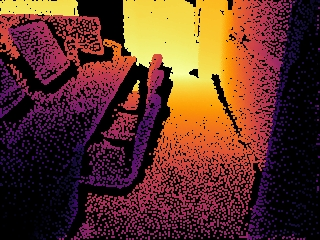}&
\includegraphics[width=2.1cm]{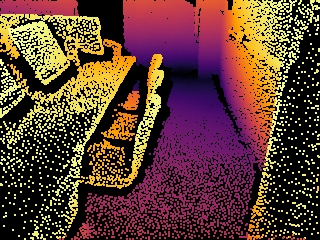}&
\includegraphics[width=2.1cm]{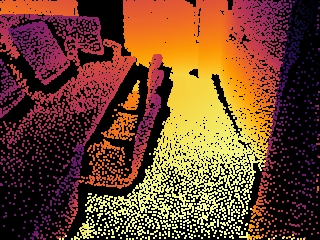}\\
(a)RGB&(b)25th channel&(c)35th channel&(d)24th channel\\
\end{tabular}
\caption{\update{LPointNet Feature Maps. The different channels of the feature maps highlight various object properties, which explains why they help the 2D semantic segmentation.}}
\label{fig:ptnetImg}
\end{figure}

\begin{figure}[!h]
\centering
\begin{tabular}{@{ }>{\centering\arraybackslash}p{2.6cm}@{ }>{\centering\arraybackslash}p{2.6cm}@{ }>{\centering\arraybackslash}p{2.6cm}}
\includegraphics[width=2.6cm]{image/ptnetIMG/scene0095_00_000600_rgb.png}&
\includegraphics[width=2.6cm]{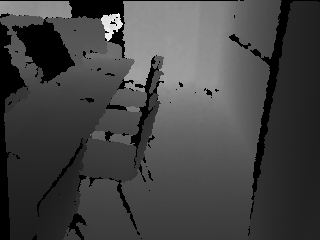}&
\includegraphics[width=2.6cm]{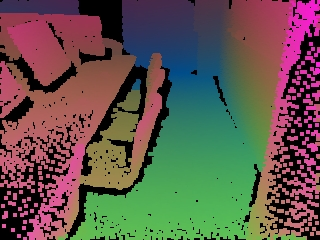}\\
(a)RGB&(b)DepthImage&(c)LPointNet Features\\
\end{tabular}
\caption{\update{LPointNet Feature Maps with PCA reduction} \update{on the number of channels for illustration purpose}.}
\label{fig:ptnetImgSCP}
\end{figure}

To illustrate the 3D features extracted by our 2D segmentation driven LPointNet, we show some channels of the features projected on the image plane (Figure \ref{fig:ptnetImg}). Some channels highlight the background (a), other channels highlight flat nearby object surfaces (b) or the floor (c), which hints at the fact that they provide important information for object segmentation.
\update{For visibility purpose, we upsample the feature map in Figure \ref{fig:ptnetImg} and \ref{fig:ptnetImgSCP} using a 2x2 \update{dilatation kernel}}

To improve the feature map visualization, we use a PCA on the 61 channel feature maps to extract the 3 most significant channels. Figure \ref{fig:ptnetImgSCP} shows the RGB, depth images and corresponding PCA feature maps.
In the image obtained, we can see that the pixel colors highlight interesting cues such as the distance to the camera, the \update{relative orientation and position} of the objects in the scene. It also shows interesting segmentation cues, giving a same color to an object.

\begin{figure}[!h]
\centering
\includegraphics[width=8.2cm]{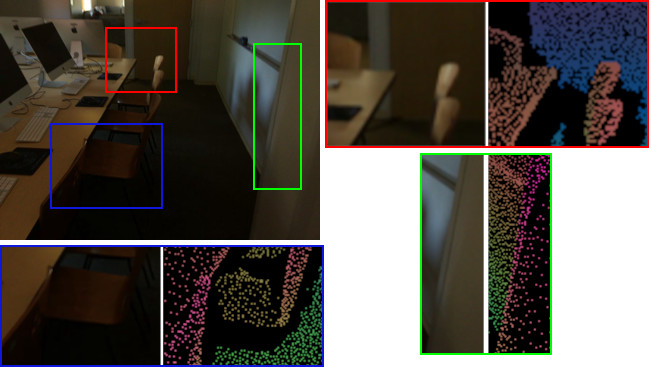}
\caption{\update{Closeups on the extracted features. LPointNet extracts relevant features} \update{on the relative position and orientation of objects.} }
\label{fig:zoomFeat}
\end{figure}

\update{Figure \ref{fig:zoomFeat} shows some closeups in areas with multiple edges and depth transition between objects. }
\update{Importantly enough for the segmentation task, } \update{one can see that the LPointNet extracted features capture and highlight such transitions when projected on the image plane. Therefore the projection preserves the objects' boundaries.}

\section{Discussion}
While our framework consistently improves 2D segmentation network performances without requiring additional groundtruth labels, some much heavier methods (VirtualMVFusion~\citep{kundu2020mv}) yield better mIoU performances, at the cost of requiring either 3D labels, multi-view images, synthetic data augmentation or a heavier network.
Overall, our network exhibits good quantitative performances measured by the mIoU metric (mean Intersection over Union) which is the standard quality measure for segmentation tasks. However, Figure \ref{fig:segImg1} illustrates that this ground truth segmentation is sometimes wrong. For example, some parts have no labels while they should. Our method, aided by 3D geometric features, can better recover the labels in these missing areas, which is good in terms of real performance but is not always reflected in the measured mIoU.

\section{Conclusion}

In this paper we introduced a 2D segmentation method taking advantage of 3D geometric data without needing explicit 3D network supervision.
Our method yields competitive segmentation results, outperforming other segmentation methods, with a relatively small cost compared to recent approaches.
While we have used our approach for the segmentation of individual images, it is clear that the performances would be further improved by using multiple images, as cross-image label consistency could provide a way to discard wrong labels, a direction we will explore in a future work.

\section{Acknowledgments}
The authors acknowledges support from ANRT, PhD grant n◦2019/1705.
This work was granted access to the HPC/AI resources of IDRIS under the allocation 2021-AD011012380  made by GENCI.

\bibliographystyle{cag-num-names}
\bibliography{refs}

\appendix
\section{Data preparation} \label{app:dataset}

\subsection{ScanNet dataset}
The ScanNet dataset \cite{dai2017scannet} provides images of various scenes, acquired by a RGBD capture system, together with camera poses and the scene's point cloud.
In our case, we use only the RGB frames and the point cloud consolidated from the depth images.

The color images are shot as a video, and are therefore redundant. The amount of images is also expensive in memory which has a strong impact on the training time. To alleviate this issue, we take only an image every $20$ frames, resulting in a $95$k images subset for the training set.

In addition, some camera poses given in the dataset are corrupted, with invalid values of camera's parameters, we removed the corresponding frames because our network relies on the correct setting of the camera's field of view to merge pixels and points features.
The ScanNet dataset provides two pointclouds, one dense and one sparse.
In order to reduce the training time and memory consumption, we trained the network using the sparse version of the point cloud.

\subsection{2D-3D-S dataset}
The 2D-3D-S dataset ~\citep{armeni2017dataset} provides images of offices and educational buildings for a total of 6 scenes. This dataset contains 2D, 2.5D and 3D data captured with RGBD device (Matterport camera).
We followed the guideline proposed by the authors and selected the area 1,2,3,4,6 for training, and the area 5 for testing.
Due to the number of point present in each area, and to avoid memory overflow, we downsampled the point cloud using Poisson sampling~\citep{Cook86} with a ball query of 3 cm.
As the scene covers the whole floor, we truncated the camera's viewing cone
using the max distance present in the depth image to avoid irrelevant point compared to RGB information (points coming from another room, etc.).

\subsection{NYU-V2 dataset}
The NYU-V2 dataset ~\citep{silbermanECCV12nyu} provides RGBD images coming from video captured with the Microsoft Kinect from various indoor scene.
As this dataset does not provide any 3D data, we use the camera's parameters and the depth to produce depth point cloud per image. We use directly the raw depth, as the in-painted one create 3D artifact when projected.

\section{Visibility threshold} \label{visibility}

We analyze the effect of the visibility angle parameter in the method of Pintus et al.~\citep{pintus2011visible}. All these experiments are run without any data augmentation.

\begin{table}[!htb]
\caption{Visibility angle comparison on the ScanNet~\citep{dai2017scannet} dataset. A lower angle value mean that more point will be kept by the filter. An angle with a 0 value mean that the nearest point at each pixel will be kept, independently from their 3D position in the scene.}
\centering
\begin{tabular}{p{3.8cm}|>{\centering\arraybackslash}p{3.cm}|>{\centering\arraybackslash}p{1.2cm}}
  Visibility angle value ($sr$) & Point-Pixel coverage &mIoU \\
\midrule
  angle = 0 & 21.5\% & 62.3 \\
  \textbf{angle = 2} & 15\% &\textbf{65.4} \\
  angle = 3 & 11.8\% & 65.1 \\
  angle = 4 & 8.3\% & 64.8 \\
  angle = 5 & 6.2\% & 64.3 \\
\bottomrule
\end{tabular}
\label{tab:visibilityAngle}
\end{table}

Table \ref{tab:visibilityAngle} shows that using the visibility filter impact significantly the performance of our approach when constructing the 2D geometric feature map.

\section{3D coordinate system} \label{coordinate}

PointNet~\citep{qi2016ptnet} use spatial transformers to set the points in a canonical position, these spatial transformers being themselves PointNet instances. In our case, we use the camera's parameters to set the points in the camera's coordinate system. Table \ref{tab:coordsys} shows that defining the points coordinates from the camera viewpoint helps the network to extract more relevant features compared to the world coordinate system.
We remove the data augmentation used during training to measure the validity of this choice.

\begin{table}[!htb]
\caption{Variation on the point's coordinate system on the ScanNet validation set.}
\label{tab:coordsys}
\centering
\begin{tabular}{p{6.cm}|p{2.cm}}
  Coordinate system  & mIoU \\
\midrule
  World's system per scene  &   63.8 \\
  \textbf{Camera's system}  &  \textbf{65.4} \\
\bottomrule
\end{tabular}
\end{table}

\section{Merging strategy}
We analyse the efficiency of our procedure for merging the 3D features in the 2D image analysis network. We first compare merging such features before the 2D encoder or at the end of the 2D decoder. As illustrated in Table \ref{tab:mergeFeat}, we found that merging the 3D features at an early stage allows the 2D network to better integrate 3D features. Since our 3D network's weights update comes from the 2D groundtruth, the early merge also helps the 3D network to produce more relevant features for the 2D segmentation task. In these experiments, we remove the data augmentation techniques.

To merge the RGB images and the point features images we use two different convolutional layers depending on the availability of a projected 3D information on the pixel. If no point projects on the pixel, we us a 1x1 convolution to transform it from 3 channels to 64 channels. Otherwise, we concatenate the $61$-channel geometric feature with the $3$-channel RGB feature and then use a 1x1 convolution to recombine the 64 channels in an optimal way.
Another option is to use a single 1x1 convolution operator, independently of the availability of a projected 3D point. In that case we use 0-padding for building a 64 channel descriptor of the projection-less pixels. We call this second option "Merge with padding". Table \ref{tab:mergeFeat} shows that this second option is slightly less efficient.

\begin{table}[!htb]
\caption{Merging 3D to 2D features using our PointNet variation and a 2D ResNet-34 U-Net network on the ScanNet~\citep{dai2017scannet} validation set.}
\centering
\begin{tabular}{p{7.cm}|p{1.cm}}
  Methods  & mIoU \\
\midrule
  Late merge (after 2D decoder) &  58.1 \\
  \textbf{Early merge (before 2D encoder) (Ours)} &  \textbf{65.4} \\
\midrule
  Merge with padding & 65.2 \\  
  \textbf{Local merge (Ours)} &  \textbf{65.4}  \\
\bottomrule
\end{tabular}
\label{tab:mergeFeat}
\end{table}

\section{Point cloud visibility} \label{app:pvis}
To test the importance of taking into account the visibility, we trained and tested our network on the visible point cloud, containing only the points visible from the camera (removing occluded points), and then the viewing cone of the camera, containing all points in the camera's field of view.
Importantly enough, even if LPointNet takes as input the field of view points, when projecting the features onto an image at the end of the LPointNet feature extraction, we only retain the features of the visible points, to create the features map.

Table \ref{tab:inputcloud} shows that using the field of view points instead of the visible points allows to improve the performance of the network.
A reason for this increased performance is that using the field of view point cloud gives stronger characteristics since the points represent the more global context of the images in the 3D scene, rather than just the visible part.

\begin{table}[!htb]
\caption{Visibility comparison. Using the camera's viewing cone point cloud clearly improves the performances compared to restricting to the visible point cloud. The mIoUs are given for the ScanNet~\citep{dai2017scannet} validation set.}
\centering
\begin{tabular}{p{6.cm}|p{2.cm}}
  3D input  & mIoU \\
\midrule
  Visible point cloud & 63.2 \\
  \textbf{FoV point cloud (Ours)}  & \textbf{65.4}  \\
\bottomrule
\end{tabular}
\label{tab:inputcloud}
\end{table}

\section{Kitti-360} \label{kitti-results}

We test our approach on an outdoor dataset, which poses different challenges than indoor scenes, since the field of view can be much larger (possibly infinite). We used the Kitti-360~\citep{liao2021kitti360} dataset. This dataset presents outdoors data captured using fisheye cameras alongside a laser scanning device mounted on a car. The dataset covers a driving distance of 73km with manually annotated images (19 classes).
The fisheye images are then rectified yielding a set of registered perspective images.
We train and evaluate our approach using the set of perspective images and the point cloud captured at the same time.
The dataset is split into a training set of 49k images and a validation set of 12k images. The training and validation set cover different areas of the town.

As the field of view is much larger, capturing a far larger amount of points, we adapt our projection strategy.
We first compute the visible points using the camera parameters and the visibility filter similarly to the data preparation procedure of Section \ref{data}.
We then use a K-NN radius search to gather contextual points around the visible ones.
In practice, we first downsample the point cloud using Poisson sampling~\citep{Cook86} with a ball query of 20cm.
The K-NN radius search takes points at 1 meter around the visible points.

Table \ref{tab:kittival} presents the results of our approach compared to our 2D baseline.

\begin{table}[!htb]
\caption{2D mIoU on the KITTI-360 validation set. Following the recommended procedure we removed two classes for the evaluation.}
\centering
\begin{tabular}{p{6.cm}|p{2.cm}}
  Methods  & mIoU \\
\midrule
  U-Net34 \citep{ronneberger2015unet} & 53.4 \\
  \textbf{LPointNet + U-Net34 (Ours)}  & \textbf{57.5}  \\
\bottomrule
\end{tabular}
\label{tab:kittival}
\end{table}

Compared to the indoor datasets, the mIoU is still clearly improved but the gain is smaller.
Indeed, the large field of view combined with visibility and neighborhood selection yield a point set which is much more disconnected than the ones for the indoor scenes. More precisely, it gives a set of clusters of points distant from each other. This difference in point distribution might require some network adaptation, but this deserves further investigations.

The Kitti360 is a relatively new dataset, the associated benchmark shows two 2D-only methods giving good performances, including a VGG16-FCN~\citep{long2015fcn} with a mIoU score of $54\%$, the other network is attention-based (PSPNet~\citep{zhao2017pspnet}) and reaches $64\%$ as mIoU.
These scores are only given as a guide since they are provided on the test set, while we performed our experiments on the validation set (access to the test set requiring a heavy procedure).
VGG16-FCN seems to yield a comparable (or a little smaller) score to our method, but it is much heavier (134M vs 26M parameters in our case).

The attention-based network clearly outperforms our method, but this result was expected. Networks using attention mechanisms give better results on outdoor datasets due to the object scales. A same object can indeed appear at very different scales in outdoor datasets as a consequence of the large depth field captured by the RGB images, a scale discrepancy that is very well handled by attention mechanisms.

\end{document}